# Adaptive Regularization of Ill-Posed Problems: Application to Non-rigid Image Registration

**Andriy Myronenko and Xubo Song**
Department of Science and Engineering, School of Medicine
Oregon Health and Science University, Portland, OR
`Matlab code is at www.bme.ogi.edu/~myron/matlab/MIRT.`

## Abstract

We introduce an adaptive regularization approach. In contrast to conventional Tikhonov regularization, which specifies a fixed regularization operator, we estimate it simultaneously with parameters. From a Bayesian perspective we estimate the prior distribution on parameters assuming that it is close to some given model distribution. We constrain the prior distribution to be a Gauss-Markov random field (GMRF), which allows us to solve for the prior distribution analytically and provides a fast optimization algorithm. We apply our approach to non-rigid image registration to estimate the spatial transformation between two images. Our evaluation shows that the adaptive regularization approach significantly outperforms standard variational methods.

## 1 Introduction

Tikhonov regularization has been a standard tool to tackle ill-posed problems [1, 2]. One often minimizes an objective function regularized with a smoothness constraint.

$$E(\mathbf{u}) = D(O, \mathbf{u}) + w \|\mathbf{Pu}\|^2 \tag{1}$$

where $D(O, \mathbf{u})$ is a measure of how well the solution $\mathbf{u}$ fits the given data $O$ and $\|\mathbf{Pu}\|^2$ is a regularization term that penalizes some properties of $\mathbf{u}$ (e.g. lack of smoothness, when $\mathbf{P}$ is a derivative operator). Parameter $w$ is a trade-off between data fitness and regularization. General non-quadratic forms of the regularization term have also been used.

Optimization of the regularized objective function is often challenging, because the algorithms tend to get stuck in local minima. To overcome this problem one can adjust the value of the regularization parameter $w$. Large values of $w$ allow to overcome some local minima, but result in overconstrained solutions. Thus, $w$ has to be chosen big enough to avoid local minima, but small enough to allow flexibility on $\mathbf{u}$. Multiple strategies to select $w$ have been proposed, including various heuristics, slow annealing, cross validation and Bayesian estimation [2, 3]. In many cases, there may not be a single $w$ adequate to achieve a reasonable solution.

Instead of searching for an optimal $w$ with a fixed regularization operator, we estimate the regularization operator $\mathbf{P}$, treating it as an unknown parameter. We first consider the regularization framework from a Bayesian perspective, where the regularization term comes from a Gaussian prior on $\mathbf{u}$, and $\mathbf{P}^T\mathbf{P}$ is the inverse covariance matrix (or potential matrix). Instead of fixing the prior distribution (equivalent to fixing $\mathbf{P}$), we estimate it assuming that it is close to some given model distribution. As we shall show, this allows flexibility on the prior distribution and leads to adaptive regularization. We constrain the prior distribution to be a Gauss-Markov random field (GMRF) due to the following reasons: a) a GMRF on a finite lattice has a shift invariant covariance, which allows us to solve for the covariance matrix analytically, b) the shift invariance property is consistent with derivative based regularization, and c) the known eigenstructure of the covariance matrix allows fast optimization.

We introduce our new regularization approach from a general Bayesian perspective and then consider in detail the specific problem of non-rigid image registration. In non-rigid image registration one needs to find a non-rigid transformation that aligns two given images. Non-rigid image registration is one of the key problems in computer vision with multiple application including motion correction, cross modality image fusion and atlas constraction [4, 5]. The rest of the paper is organized as follows. In Sec 2 we define a general adaptive regularization framework and describe the fast optimization algorithm. In Sec. 3 we overview the non-rigid image registration problem and show how to apply the adaptive regularization approach. In Sec. 4 we evaluate our algorithm. In Sec. 5 we compare our algorithm to related methods. Sec. 6 concludes the work with discussions.

## 2 Method

### 2.1 Bayesian formulation

From a Bayesian perspective the regularization approach is equivalent to the maximum a posteriori (MAP) estimation, that is to maximize

$$\max p(\mathbf{u}|O) \propto p(O|\mathbf{u})p(\mathbf{u}) \tag{2}$$

or equivalently to minimize the following objective function

$$\min E(\mathbf{u}) = -\log p(O|\mathbf{u}) - \log p(\mathbf{u}) \tag{3}$$

where the first term (the negative log-likelihood) is the error function $D(O, \mathbf{u})$ and $p(\mathbf{u})$ is a prior distribution on $\mathbf{u}$. In case of the quadratic regularization term (our case), the prior $p(\mathbf{u})$ is a Gaussian distribution

$$p(\mathbf{u}) = \frac{1}{\sqrt{(2\pi)^D \det(\Sigma)}} e^{-\frac{1}{2}\mathbf{u}^T \Sigma^{-1} \mathbf{u}} \tag{4}$$

Defining the inverse covariance matrix (also called potential matrix) as $\Sigma^{-1} = \mathbf{P}^T \mathbf{P}$ and substituting $\mathbf{p}(\mathbf{u})$ in Eq. 3, we achieve

$$\min E(\mathbf{u}, \mathbf{P}) = \frac{1}{w} D(O, \mathbf{u}) + \frac{1}{2} \|\mathbf{P}\mathbf{u}\|^2 - \frac{1}{2} \log \det(\mathbf{P}^T \mathbf{P}) + \frac{D}{2} \log(2\pi) \tag{5}$$

The last two terms are constants if $\mathbf{P}$ is fixed. The weight $w$ here comes from the likelihood function (this is equivalent to the regularization weight in Eq. 1). We prefer to use the covariance form instead of the operator form, and rewrite Eq. 5 as

$$\min E(\mathbf{u}, \Sigma^{-1}) = \frac{1}{w} D(O, \mathbf{u}) + \frac{1}{2} \mathbf{u}^T \Sigma^{-1} \mathbf{u} - \frac{1}{2} \log \det(\Sigma^{-1}) + const \tag{6}$$

So far we have just reformulated the regularization approach from a Bayesian perspective without any modifications. Now, instead of assuming a fixed inverse covariance matrix $\Sigma^{-1}$ (equivalent to assuming a fixed operator $\mathbf{P}$), we shall estimate it.

### 2.2 Adaptive regularization

We assume that the prior distribution $p(\mathbf{u})$ is unknown, but is close to the *model distribution* $q(\mathbf{u})$ in terms of Kullback-Leibler (KL) divergence. We rewrite the MAP problem as

$$\min E(\mathbf{u}, p) = -\log p(O|\mathbf{u}) - \log p(\mathbf{u}) + KL(p\|q) \tag{7}$$

where $KL(p\|q)$ is the KL-divergence between the unknown prior distribution $p$ and the model distribution $q$

$$KL(p\|q) = \int p(\mathbf{u}) \log \frac{p(\mathbf{u})}{q(\mathbf{u})} d\mathbf{u} \tag{8}$$

We want to minimize Eq. 7 simultaneously with respect to $\mathbf{u}$ and $p$. The prior distribution $p$ is multidimensional in general. To simplify and stay consistent with conventional regularization, we assume that $p$ and $q$ are zero-mean multivariate Gaussian with covariances $\Sigma$ and $\Omega$ respectively:

$p \sim \mathcal{N}(0, \Sigma), q \sim \mathcal{N}(0, \Omega)$. One advantage of such assumption is that the KL-divergence has an analytical form

$$KL(p||q) = \frac{1}{2} \left( \text{tr}(\Omega^{-1}\Sigma) - \log \det(\Omega^{-1}) + \log \det(\Sigma^{-1}) - N \right) \tag{9}$$

which we substitute in Eq. 7 to obtain

$$\min E(\mathbf{u}, \Sigma^{-1}) = \frac{1}{w} D(O, \mathbf{u}) + \frac{1}{2} \mathbf{u}^T \Sigma^{-1} \mathbf{u} + \frac{1}{2} \text{tr}(\Omega^{-1}\Sigma) + const \tag{10}$$

We only need to find $\Sigma$ to fully specify the prior distribution $p$. We can analytically solve for $\Sigma$ in at least two cases: a) when both $\Sigma$ and $\Omega$ are diagonal, b) when $\Sigma$ and $\Omega$ commute (share the eigenvectors). Indeed the second case arises naturally in regularization theory.

### 2.3 Markov random field and shift invariance

We constrain the prior distribution to be a Gauss-Markov random field (GMRF), which implies that the inverse covariance matrix is shift invariant[1] and is diagonalized by the Fourier basis on the discrete lattice [6]. Matrices $\Sigma^{-1}$ and $\Omega^{-1}$ are symmetric positive definite by definition and allow spectral decomposition

$$\Sigma^{-1} = \mathbf{Q}\Lambda\mathbf{Q}^T, \quad \Omega^{-1} = \mathbf{Q}\mathbf{K}\mathbf{Q}^T \tag{11}$$

where $\Lambda = \text{d}[\lambda_1, .., \lambda_N], \forall \lambda_n \geq 0$ is a diagonal matrix of the eigenvalues of $\Sigma^{-1}$ and $\mathbf{K} = \text{d}[k_1, .., k_N], \forall k_n \geq 0$, is a diagonal matrix of the eigenvalues of $\Omega^{-1}$. For the shift invariant matrix, the eigenvectors $\mathbf{Q} = [\mathbf{q}_1, .., \mathbf{q}_N]$ have a known form. Depending on the boundary conditions, $\mathbf{Q}$ is a discrete Fourier basis (circular boundary conditions) or discrete cosine transform (DCT) basis (Neumann boundary conditions) [6, 7]. The assumption of the GMRF structure has multiple advantages. First, estimation of the inverse covariance matrix simplifies to estimation of its eigenvalues. Second, the product $\mathbf{Q}^T\mathbf{x}$ is a multidimensional DFT (or DCT) transform and can be computed fast in $\mathcal{O}(N \log N)$. Finally, the shift invariant structure of the covariance matrix is consistent with standard derivative-based regularization operators.

Common regularization operators $\mathbf{P}$, such as first, second, and higher order derivative operators are shift invariant by definition. In the discrete case, this means that the matrix

$$\Sigma^{-1} = \mathbf{P}^T\mathbf{P} \tag{12}$$

is a Toeplitz plus nearly Hankel matrix [7, 8]. Such a highly structured matrix is known to be diagonalized by a Fourier basis, and the estimation of $\mathbf{P}^T\mathbf{P}$ reduces to an estimation of its eigenvalues [9]. Substituting Eq. 11 into Eq. 10, we obtain

$$\min E(\mathbf{u}, \Lambda) = \frac{1}{w} D(O, \mathbf{u}) + \frac{1}{2} \sum \lambda_i (\mathbf{q}_i^T \mathbf{u})^2 + \frac{1}{2} \sum \frac{k_i}{\lambda_i} \tag{13}$$

We equate the gradient of the function to zero, and solve for $\Lambda$ to obtain:

$$\lambda_i = \frac{\sqrt{k_i}}{|\mathbf{q}_i^T \mathbf{u}|} \tag{14}$$

where $|\cdot|$ denotes the absolute value. The solution for the eigenvalues $\Lambda$ is guaranteed to be positive. Substituting $\Lambda$ back into Eq. 13, we obtain

$$\min E(\mathbf{u}) = \frac{1}{w} D(O, \mathbf{u}) + \sum \sqrt{k_i} |\mathbf{q}_i^T \mathbf{u}| \tag{15}$$

At this point, we have analytically solved for $\Lambda$ (which uniquely specifies the prior distribution $p$) and eliminated it from the objective function. The final form of the objective function includes the regularizer that penalizes the absolute value of Fourier (or DCT) coefficients of $\mathbf{u}$ weighted by $K$. We still need to specify the eigenvalues $K$ of the model distribution $q$.

To choose the model distribution $q$, and in particular its inverse covariance matrix $\Omega^{-1}$, we follow a standard regularization approach. For instance, Laplacian $\|\Delta \mathbf{u}\|^2$ is a popular regularizer, which suggests to use $\Omega^{-1} = \Delta^2$ as the inverse covariance matrix of the model distribution. The eigenvalues of the discrete Laplacian (squared) on a regular grid (with Neumann boundary condition [7]) in 1D are $\mathbf{k}^2 = [k_1^2, ..k_N^2]^T$, $k_n = 2(1 - cos(\pi(n-1)/N))$, $n = 1, .., N$.

---

[1] By shift invariant matrix, we denote a Toeplitz plus nearly Hankel matrix (due to the boundary conditions).

## 2.4 Optimization

To optimize the objective function in Eq. 15, we take advantage of the fact that the original objective function (Eq. 10) had a quadratic regularization term, and split the optimization into two steps:

$$\bullet \text{ find } \Sigma^{-1}: \quad \Lambda = \mathbf{K}^{1/2} \, diag(|\mathbf{Q}^T \mathbf{u}|)^{-1}, \Sigma^{-1} = \mathbf{Q}\Lambda\mathbf{Q}^T, \tag{16}$$

$$\bullet \text{ minimize } E(\mathbf{u}) = D(O, \mathbf{u}) + w\mathbf{u}^T \Sigma^{-1} \mathbf{u} \tag{17}$$

The first step, solution for $\Sigma^{-1}$, has a closed form, whereas the second step requires iterative minimization (unless $D(O, \mathbf{u})$ is quadratic). We shall briefly state one of the standard fast optimization approaches to minimize the function with quadratic penalty term.

We equate the gradient of the objective function in Eq. 17 to zero

$$\nabla E(\mathbf{u}) = \nabla D(O, \mathbf{u}) + w\Sigma^{-1}\mathbf{u} = 0 \tag{18}$$

Note that the gradient consists of the non-linear part ($\nabla D(O, \mathbf{u})$) and the linear part ($\Sigma^{-1}\mathbf{u}$). We artificially introduce a time-step derivative to the right-hand side of Eq. 18 as

$$\nabla D(O, \mathbf{u}^t) + w\Sigma^{-1}\mathbf{u}^{t+1} = -(\mathbf{u}^{t+1} - \mathbf{u}^t)/\gamma \tag{19}$$

which converges to zero in equilibrium. Here $\gamma$ is a time step parameter (similar to the gradient descent step size parameter). Note that the linear part is at time $t+1$, whereas the non-linear part is kept at $t$. Solving for $\mathbf{u}^{t+1}$, we achieve

$$\mathbf{u}^{t+1} = \mathbf{Q}(\mathbf{I} + \gamma w\Lambda)^{-1}\mathbf{Q}^T(\mathbf{u}^t - \gamma \nabla D(O, \mathbf{u}^t)) \tag{20}$$

Iterating on $\mathbf{u}$ we achieve faster minimization compared to the first-order minimization methods, with no need for Hessian computations or approximations. The eigenvector matrix $\mathbf{Q}$ is never constructed explicitly. The matrix vector products $\mathbf{Q}^T\mathbf{x}$ and $\mathbf{Q}\mathbf{x}$ are forward and inverse multidimensional DCTs respectively. Finally, combining Eq. 16 and Eq. 20 into a single step, we achieve our fast optimization algorithm by iterating on

$$\mathbf{u}^{t+1} = \mathbf{Q}\frac{\mathrm{d}\,|\mathbf{Q}^T\mathbf{u}^t|}{(\mathrm{d}\,|\mathbf{Q}^T\mathbf{u}^t| + \gamma w\mathbf{K})}\mathbf{Q}^T(\mathbf{u}^t - \gamma \nabla D(O, \mathbf{u}^t)) \tag{21}$$

where $\mathrm{d}\,|\cdot|$ denotes a diagonal matrix formed from the right-hand side vector.

## 3 Non-rigid Image Registration

Image registration is one of the key problems in computer vision. The goal of image registration is to find a spatial transformation that aligns two images. The most challenging cases of image registration occur when the underlying transformation is non-rigid. Non-rigid image registration has a wide range of applications, including motion correction, change detection over time, cross modality image fusion, inter-subject comparison, atlas constraction, registration-based segmentation, motion estimation and tracking [4, 5].

As far as non-rigid transformation is a broad class of nonlinear transformations, non-rigid image registration is an ill-posed problem. To tackle the problem one can use either parametric or non-parametric (also called variational) approaches. Parametric image registration specifies a parametric model of the transformation (e.g., locally affine or B-splines), which explicitly constrains the transformation [10]. This, however, significantly limits the range of admissible transformations, and is not adequate for complex non-rigid transformations. Non-parametric image registration estimates the transforamtion as an unknown function using variational calculus. One often uses a regularizer on the transformation to make the problem well-posed.

One of the standard non-parametric approach is to minimize the following objective function [11, 12]

$$E(\mathbf{u}) = D(I, J_{\mathbf{u}}; \mathbf{u}) + w \left\|\Delta \mathbf{u}\right\|^2 \tag{22}$$

where $D(I, J_{\mathbf{u}}; \mathbf{u})$ is a similarity measure between the images $I$ and $J$, and $\mathbf{u}$ is a displacement field that aligns $J$ onto $I$. Regularization of the displacement function is often called the competitive regularization, in contrast to the incremental regularization, which penalize increments (or evolution) of the displacement function [13]. Unfortunately, regularization of the displacement function (and its increments) do not guarantee a diffeomorphic transformation. A diffeomorphic transformation ensures that an inverse transformation and its derivatives exist and are smooth functions, which can be required in medical images. In a few cases when the standard regularization does not produce diffeomorphic transformation, it can be post-smoothed to ensure the invertibility [14]. A more elegant approach to ensure diffeomorphism is to consider the transformation as a solution of the ordinary differential equation [15, 16]. Such diffeomorphic image registration methods explicitly account for invertible transformation, but suffer from large computational complexity. The detailed overview of the non-rigid transformation models in image registration is beyond the scope of this paper. Here, we apply our regularization approach to estimate the displacement function, which is most similar in formulation to the competitive regularization in Eq. 22.

**Adaptive regularization:** Following our adaptive regularization method, we minimize the following objective function

$$\min E(\mathbf{u}) = \frac{1}{w} D(I, J_\mathbf{u}; \mathbf{u}) + \sum \sqrt{k_i} |\mathbf{q}_i^T \mathbf{u}| \tag{23}$$

where $D(I, J_\mathbf{u}; \mathbf{u})$ is a similarity measure between the images $I$ and $J$. We shall use the sum-of-squared-differences (SSD) similarity measure, defined as $D(I, J_\mathbf{u}; \mathbf{u}) = \sum_i \left(I(\mathbf{x}_i) - J(\mathbf{x}_i + \mathbf{u}(\mathbf{x}_i))\right)^2$. The choice of the similarity measure is itself a research topic in image registration. Here, we use a simple similarity measure to concentrate primarily on the transformation estimation. Any other similarity measures can be easily applied for $D(I, J_\mathbf{u}; \mathbf{u})$. We remind that $\mathbf{q}_i$ are the DCT bases, which corresponds to the Neumann boundary condition on $\mathbf{u}$. Such condition is often the most appropriate boundary condition for natural images.

**Implementation:** To optimize the objective function in Eq. 23, we follow our iterative update according to Eq. 21. To clarify the algorithm in multidimensional cases (e.g., for 3D images), we rewrite Eq. 21 as

$$|\mathbf{Q}^T \mathbf{u}^t| \equiv \sqrt{DCT^2(\mathbf{u}_x^t) + DCT^2(\mathbf{u}_y^t) + DCT^2(\mathbf{u}_z^t) + \epsilon}$$

$$\mathbf{u}_i^{t+1} = IDCT \left( \frac{|\mathbf{Q}^T \mathbf{u}^t|}{(|\mathbf{Q}^T \mathbf{u}^t| + \gamma w \mathbf{K})} \cdot DCT(\mathbf{u}_i^t - \gamma \nabla_i D(O, \mathbf{u}^t)) \right), \text{for } i = x, y, z$$

where $DCT$ and $IDCT$ denote forward and inverse multidimensional (in this case 3D) discrete cosine transforms, and all operations are elementwise. We add a small positive constant $\epsilon$ (e.g. machine precision) to avoid devision by zero. $\mathbf{u}_x, \mathbf{u}_y$ and $\mathbf{u}_z$ are the 3D arrays of cooresponding voxel displacements. Array $\mathbf{K}$ denotes the eigenvalues of the multidimensional Laplacian with elements: $k_{ijp} = \mathcal{K}(i, N_x) + \mathcal{K}(j, N_y) + \mathcal{K}(p, N_z)$, where $\mathcal{K}(n, N) = 2(1 - cos(\frac{\pi(n-1)}{N}))$, and $N_x, N_y, N_z$ are the image dimensions. Finally for the SSD similarity measure, the gradient is $\nabla_i D(O, \mathbf{u}^t) = (J(\mathbf{x} + \mathbf{u}) - I(\mathbf{x})) \nabla_i J(\mathbf{x} + \mathbf{u})$, for $i = x, y, z$.

## 4 Results

We have implemented our algorithm in Matlab, and tested it on a AMD Opteron CPU 2GHz Linux machine with 4GB RAM. We used the BrainWeb T1-weighted MRI images ($180 \times 216 \times 180$ voxels) to test the algorithm [17]. We normalized image intensities to the $[0, 1]$ interval before registration. The stopping criterion was either when the algorithm reaches 1000 iterations or the objective function tolerance drops below $10^{-8}$.

To simulate the synthetic spatial deformation we put a uniform grid of control points (with $15\%$ spacing) over the image, randomly perturb them ($\mathcal{N}(0, \sigma = 6)$) and interpolate the image using thin plate spline [18]. This way we obtain known smooth locally varying synthetic deformations.

We compare our algorithm to the curvature based variational registration [11, 12], which has a quadratic Laplacian regularization term as in Eq. 22. To evaluate the registration performance, we compute the root mean squared error (RMSE) between the true and estimated transformations as

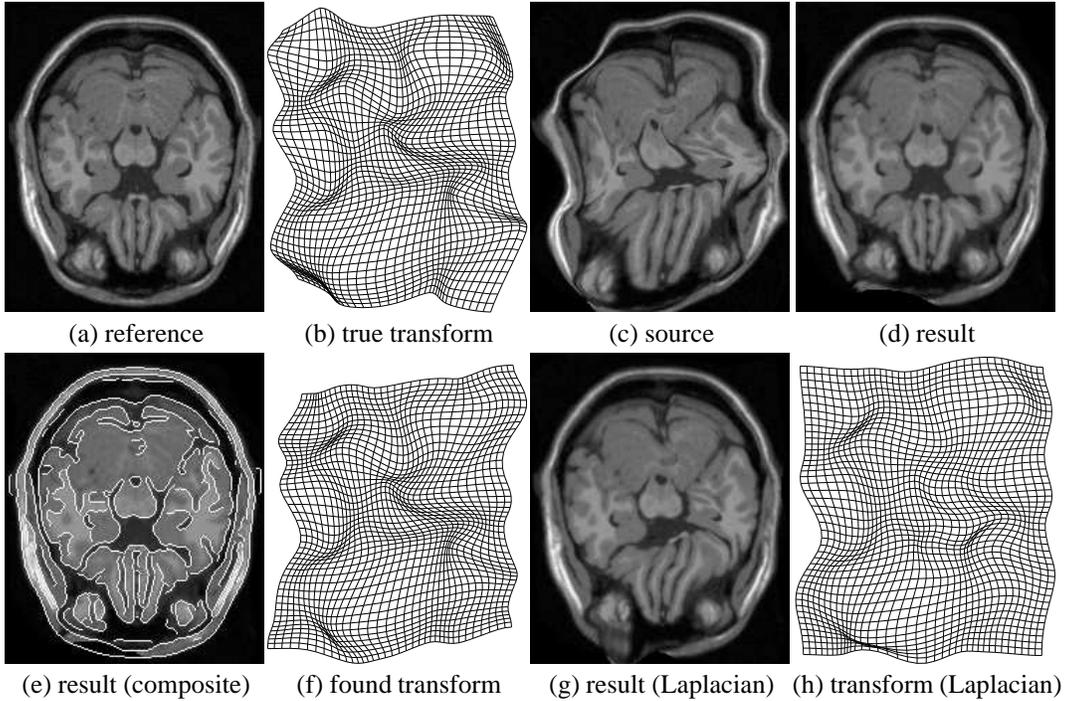

(a) reference    (b) true transform    (c) source    (d) result

(e) result (composite)    (f) found transform    (g) result (Laplacian)    (h) transform (Laplacian)

Figure 1: Non-rigid registration example with complex synthetic transformation. We register the source image (c) onto the reference image (a). The registration result (d) is accurate. Figure (e) shows the composite view of registered image and the contours extracted form the reference image. The estimated transformation (f) is very similar to the true transformation (b) with $\varepsilon_{RMSE} = 0.52$ transformation error. Note, that we do not include the area outside the scull into the error computation to avoid the boundary error influence. We compare our results to the Laplacian-based regularization (g,h), which has less satisfactory performance with $\varepsilon_{RMSE} = 2.21$.

$\varepsilon_{RMSE} = \frac{1}{3N} \sum \|\mathbf{u}_{true} - \mathbf{u}_{estimated}\|^2$. We do not include the area outside the scull (roughly found by thresholding) for the $RMSE$ computation to avoid the boundary error influence.

First, we demonstrate the registration performance on 2D images ($216 \times 180$). Figure 1(a,b,c) shows the reference image, its synthetically deformed version (source image) and the true transformation (which gives 7.32 initial RMSE). Figure 1(d,e,f) shows our registration result. Our algorithm accurately estimates the transformation with $\varepsilon_{RMSE} = 0.52$. Laplacian-based registration (g,h) produces less satisfactory result with $\varepsilon_{RMSE} = 2.21$. Notice, that the Laplacian-based approach did align some less challenging parts of the image, but got stuck in a local minima.

Figure 2 demonstrates the full volume 3D non-rigid image registration. The registration performance is accurate. We conducted several tests to evaluate and compare our approach to the standard quadratic (Laplacian-based) regularization. We performed 100 non-rigid 3D image registration experiments with random non-rigid deformation (similar to the ones in Fig. 1b) initialization at each run. The average initial transformation RMSE was 8.36. We did 20 experiments for each of the five different regularization weight values. We used $w = [0.5, 1.2, 2, 3, 4]$ for our algorithms and $w = [5, 10, 20, 30, 50]$ for Laplacian-based regularization. Such ranges were empirically found to be optimal regularization weight values for our data sets. Figure 3 shows the estimated transformation RMSE for different values of $w$. Our adaptive regularization approach is accurate with an average transformation error below 1 voxel, whereas quadratic regularization has less satisfactory performance.

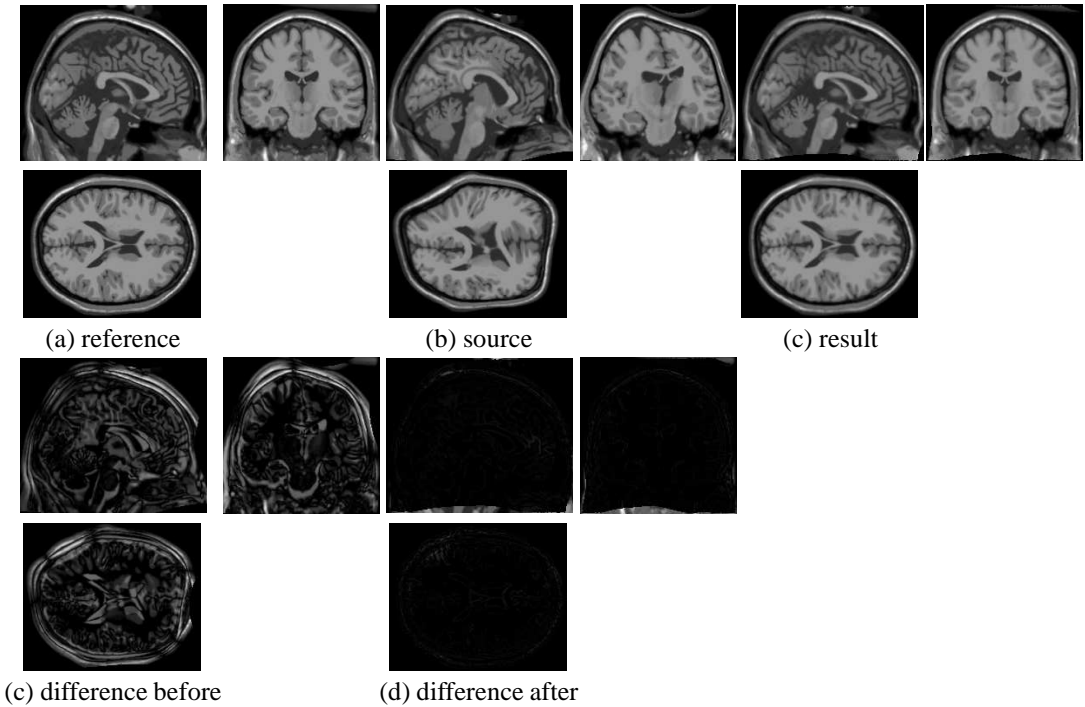

(a) reference      (b) source      (c) result

(c) difference before      (d) difference after

Figure 2: 3D non-rigid registration example. We register the source image (b) onto the reference image (a) to obtain the registered image (c). The difference image (d) between the reference and registered image is almost zero, which demonstrates the accuracy of the registration. The estimated average transformation error is $\varepsilon_{RMSE} = 0.82$.

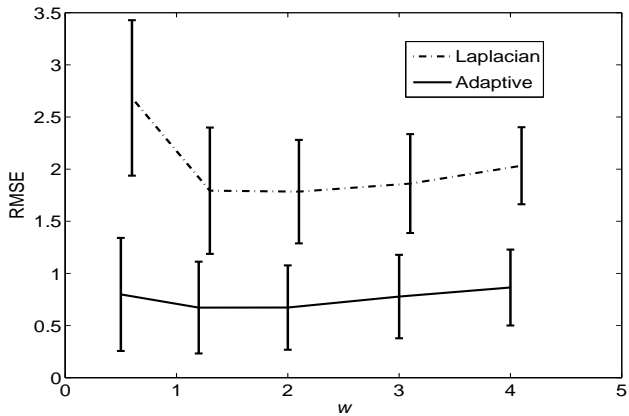

Figure 3: Comparison of the non-rigid image registration performances of our algorithm and Laplacian (curvature) based method. We conducted 20 experiments for each of the regularization weight values. We used $w = [0.5, 1.2, 2, 3, 4]$ for our algorithms and $w = [5, 10, 20, 30, 50]$ for Laplacian-based regularization. The Laplacian-based regularization results were rescaled into the common range $[0..5]$ for simpler visualization. Our adaptive regularization approach is accurate with average transformation error below 1 voxel, whereas standard Laplacian-based regularization has less satisfactory performance.

## 5 Related work

**Analogy with Kernel matrix completion:** Tsuda et al. [19] proposed a method to complete a kernel matrix with auxiliary data. In their work, the input data is available only for a subset of samples, and the kernel matrix derived from such data has missing entries. To complete the kernel matrix, the authors use an auxiliary kernel matrix derived from another information source. They minimize the Kullback-Leibler (KL) divergence between two kernel matrices and make use of the Riemannian information geometry, where the KL-divergence is defined by relating the kernel matrix to a covariance matrix of Gaussian distribution. The KL-divergence allows to use the *em* algorithm [20] (different from the EM algorithm of Dempster et al. [21]). In these terms, by minimizing $KL(p||q)$ (or in matrix notations $KL(\Sigma^{-1}||\Omega^{-1})$ in Eq. 9), we are finding (on the manifold) a positive definite matrix $\Sigma^{-1}$ that is closest to the given matrix $\Omega^{-1}$. Relating our formulation to the kernel completion one, we are simultaneously completing the inverse covariance matrix $\Sigma^{-1}$ from the current (estimated) observations $\mathbf{u}^t$ and the auxiliary matrix $\Omega^{-1}$. This allows flexibility on $\Sigma^{-1}$, in contrast to its fixed form formulation.

**Analogy with $L_1$-norm regularization:**  Consider a simple case of weighted $L_2$ norm regularization.

$$\min E(\mathbf{u}, \mathbf{a}) = \frac{1}{w} D(O, \mathbf{u}) + \sum_i a_i u_i^2 \qquad (24)$$

where $a_i \geq 0$ are unknown weights. This is equivalent to the assumption of a strict diagonal form of the covariance matrix $\Sigma$. To see the analogy, the last term here can be written as $\sum a_i u_i^2 = \mathbf{u}^T \Sigma^{-1} \mathbf{u}$, where $\Sigma^{-1}$ has elements $a_i$ along its diagonal. We assume the model distribution $q$ to be a Gaussian with *isotropic* diagonal covariance of all ones, that is $q(\mathbf{u}) \propto e^{-\|\mathbf{u}\|^2/2}$. Following our derivation we can analytically solve for $a_i$, eliminate it from the equation and achieve the following objective function

$$\min E(\mathbf{u}) = \frac{1}{w} D(O, \mathbf{u}) + \sum_i |u_i| \qquad (25)$$

which is the $L_1$ regularized problem. Thus, optimizing the error function with $L_2$ weighted norm (with unknown weights) is equivalent to the $L_1$ regularized problem (also called Lasso in regression problems [22]). This provides another interpretation of $L_1$ norm regularizers that recently attracted a lot of interests in the machine learning community [23, 24].

**Analogy with Sparseness and Compression:**  In our final objective function (Eq. 15), the last term $\sum k_i |\mathbf{q}_i^T \mathbf{u}|$ represents the $L_1$ norm of the DCT (or FFT) coefficients of $\mathbf{u}$. $L_1$ norm has been popular to enforce sparseness of the coefficients [25]. From this perspective, the standard quadratic regularizer can be seen as the one that penalizes $L_2$ norm of the DCT coefficients, e.g. $\|\Delta \mathbf{u}\|^2 = \mathbf{u}^T \mathbf{Q} \mathbf{K}^2 \mathbf{Q} \mathbf{u} = \sum k_i^2 (\mathbf{q}_i^T \mathbf{u})^2$ (compare to our $\sum k_i |\mathbf{q}_i^T \mathbf{u}|$). Thus using our regularization approach we are also enforcing sparseness of the DCT coefficients of $\mathbf{u}$. Sparseness of the estimated signal often leads to its better generalization properties [23]. As a few DCT coefficients include most of the signal information, sparseness of the DCT coefficient also forces higher compression of the estimated signal.

**Analogy with Adaptive Filtering:**  Finally, we draw the analogy of our approach with adaptive filtering. Optimization with a standard quadratic regularizer is equivalent to unregularized optimization followed by filtering, where the filter depends on the regularization operator [26]. Indeed, Eq. 20 represents a gradient descent step, followed by the fixed low-pass filter $(\mathbf{I} + \gamma w \Lambda)^{-1}$ in frequency domain. In our adaptive regularization approach, the filter becomes signal-dependent: $\frac{\mathrm{d}|\mathbf{Q}^T \mathbf{u}^t|}{(\mathrm{d}|\mathbf{Q}^T \mathbf{u}^t| + \gamma w \mathbf{K})}$ (see Eq. 21). Such a filter also resembles the Wiener filter, where the power spectrum of the actual signal is $|\mathbf{Q}^T \mathbf{u}^t|$ (signal from the previous iteration) and the noise power spectrum is $\gamma w \mathbf{K}$. The important fact here is that the filter is adaptive, it changes at every iteration, whereas for the standard regularization operator, the filter is fixed.

## 6  Discussion and Conclusion

We introduced adaptive regularization approach. Instead of assuming a fixed regularization operator (or a fixed prior distribution on the function/parameters), we estimate it. We assume the prior distribution on parameters to be close to the given model distribution in terms of KL-divergence. We constrain the prior distribution to be a Gauss-Markov random field, which allows us to solve for the prior distribution analytically and eliminate it from the equation. The final objective function appears to have a regularization term that penalizes the absolute value of **u** after applying an orthogonal transformation (Fourier or DCT). DCT approximates the optimal (in the decorrelation sense) Karhunen-Loève transform with certain Markovian assumptions [8, 7]. Penalizing the absolute value of the decorelated vector **u**, we are also enforcing sparseness on **u** in terms of the basis functions (DCT, DFT), which leads to better compression and generalization properties. We also proposed the fast optimization algorithm with complexity of $\mathcal{O}(N \log N)$.

Using our regularization approach, we achieved accurate non-rigid image registration results. Our method recovered challenging non-rigid transformation fields, whereas standard variational methods had less satisfactory results and usually converged to a "bad" local minimum. We still have to choose the regularization weight and the regularization operator (covariance matrix of the model distribution) similar to the standard quadratic regularization. Nevertheless, our approach does not constrain the regularization operator to a fixed form and allows it to be flexible.

## References

[1] A. N. Tikhonov and V. I. Arsenin. *Solutions of Ill-Posed Problems*. Winston and Sons, 1977.

[2] Z. Chen and S. Haykin. On different facets of regularization theory. *Neural Computation*, 14(12):2791–2846, 2002.

[3] C.J. Yang, R. Duraiswami, and L.S. Davis. Near-optimal regularization parameters for applications in computer vision. In *ICPR*, pages 569–573, 2002.

[4] W. R. Crum, T. Hartkens, and Derek L. G. Hill. Non-rigid image registration: Theory and practice. *The British Journal of Radiology*, 77(special issue):S140–S153, December 2004.

[5] D. L. G. Hill, P. G. Batchelor, M. Holden, and D. J. Hawkes. Medical image registration. *Physics in medicine and biology*, 46(3):R1–R45, March 2001.

[6] J.M.F. Moura and M.G.S. Bruno. DCT/DST and Gauss-Markov fields: conditions for equivalence. *Signal Processing, IEEE Transactions on*, 46(9):2571–2574, 1998.

[7] Gilbert Strang. The discrete cosine transform. *SIAM Review*, 41(1):135–147, 1999.

[8] V. Sanchez, P. Garcia, A. M. Peinado, J. C. Segura, and A. J. Rubio. Diagonalizing properties of the DCT. *IEEE Trans. Signal Processing*, 43(11), 1995.

[9] Jorge Larrey-Ruiz, Rafael Verdú-Monedero, and Juan Morales-Sánchez. A fourier domain framework for variational image registration. *J. Math. Imaging and Vision*, 32(1):57–72, 2008.

[10] D. Rueckert, L.I. Sonoda, C. Hayes, D.L.G. Hill, M.O. Leach, and D.J. Hawkes. Nonrigid registration using free-form deformations: application to breast mr images. *Medical Imaging, IEEE Transactions on*, 18(8):712–721, 1999.

[11] Jan Modersitzki. *Numerical Methods for Image Registration*. Oxford university press, 2004.

[12] B. Fischer and J. Modersitzki. A unified approach to fast image registration and a new curvature based registration technique. *Linear Algebra and its Applications*, 380:107–124, 2004.

[13] Pascal Cachier, Eric Bardinet, Didier Dormont, Xavier Pennec, and Nicholas Ayache. Iconic feature based nonrigid registration: the pasha algorithm. *CVIU*, 89(2-3):272–298, 2003.

[14] Ardekani B.A., Guckemus S., Bachman A., Hoptman M.J., Wojtaszek M., and Nierenberg J. Quantitative comparison of algorithms for inter-subject registration of 3d volumetric brain mri scans. *J. Neurosci. Methods*, 142(1):67–76, 2005.

[15] M. F. Beg, M. I. Miller, A. Trouvé, and L. Younes. Computing large deformation metric mappings via geodesic flows of diffeomorphisms. *IJCV*, 61(2):139–157, 2005.

[16] J. Ashburner. A fast diffeomorphic image registration algorithm. *NeuroIm.*, (38):95–113, 2007.

[17] C.A. Cocosco, V.Kollokian, R.K.-S. Kwan, and A.C. Evans. Brainweb: Online interface to a 3D mri simulated brain database. In *ICFMHB*, page S425, 1997.

[18] F. L. Bookstein. Principal warps: Thin-plate splines and the decomposition of deformations. *PAMI*, 11(6):567–585, 1989.

[19] Koji Tsuda, Shotaro Akaho, and Kiyoshi Asai. The em algorithm for kernel matrix completion with auxiliary data. *J. Machine Learning Research*, 4:67–81, 2003.

[20] Shun-Ichi Amari. Information geometry of the em and em algorithms for neural networks. *Neural Netw.*, 8(9):1379–1408, 1995.

[21] A. P. Dempster, N. M. Laird, and D. B. Rubin. Maximum likelihood from incomplete data via the em algorithm. *Journal of the Royal Statistical Society.*, 39(1):1–38, 1977.

[22] R. Tibshirani. Regression shrinkage and selection via the lasso. *Journal of Royal Statistical Society Series B*, 1996.

[23] Andrew Y. Ng. Feature selection, l1 vs. l2 regularization, and rotational invariance. In *ICML*, page 78. ACM, 2004.

[24] Galen Andrew and Jianfeng Gao. Scalable training of l1-regularized log-linear models. In *ICML*, pages 33–40. ACM, 2007.

[25] Bruno A. Olshausen and David J. Field. Sparse coding with an overcomplete basis set: a strategy employed by v1. *Vision Research*, 37:3311–3325, 1997.

[26] Pascal Cachier and Nicholas Ayache. Isotropic energies, filters and splines for vector field regularization. *J. Math. Imaging Vis.*, 20(3), 2004.